# Hybrid Neural Network Architecture for On-Line Learning

**Yuhua Chen**[*†] · **Subhash Kak**[‡] · **Lei Wang**[§]

**Abstract** Approaches to machine intelligence based on brain models have stressed the use of neural networks for generalization. Here we propose the use of a hybrid neural network architecture that uses two kind of neural networks simultaneously: (i) a surface learning agent that quickly adapt to new modes of operation; and, (ii) a deep learning agent that is very accurate within a specific regime of operation. The two networks of the hybrid architecture perform complementary functions that improve the overall performance. The performance of the hybrid architecture has been compared with that of back-propagation perceptrons and the CC and FC networks for chaotic time-series prediction, the CATS benchmark test, and smooth function approximation. It has been shown that the hybrid architecture provides a superior performance based on the RMS error criterion.

## 1  Introduction

On-line learning requires the learning of the mode or context, out of a set of many, within which the time-varying system is evolving. A few examples of time-varying systems are: aircraft during flight since the configurations of control surfaces as well as flight conditions and the weight of the aircraft are continually changing; the human vocal tract, as the shape of the vocal organs gets modified in the production of different vocalizations; electric or computer networks which may have different modes of behavior depending of the time of the day; and the financial system which changes according to the investor sentiment. In this paper, we consider such non-standard pattern recognition and discovery problems where statistical techniques and conventional neural networks are not convenient to use because of their slow speed.

Pattern discovery may be based on statistical methods [1], [2] or on neural networks. The currently researched techniques include various kinds of neural networks [3], principal-component analysis [4], principal-component regression

---

[*] Corresponding author, e-mail: yuhua.chen@mail.uh.edu
[†] Department of Electrical and Computer Engineering, University of Houston, Houston, TX 77204-4005, USA
[‡] Department of Computer Science, Oklahoma State University, Stillwater, OK 74078, USA
[§] Department of Electrical and Computer Engineering, University of Houston, Houston, TX 77204-4005, USA

[4], partial least-squares regression [5], and combinatorial methods [6]. The Vapnik-Chervonenkis dimension [1] quantifies the ease of learning categories from small data sets, and if the dimension is finite, machine learning of a certain kind can be efficiently done. The time-varying nature of the data sets that we wish to consider here precludes this approach. Machine intelligence that is based on brain models has stressed the use of neural networks [7] that learn from a training set and then generalize on new input data. However, it has been argued [8], [9] that such models miss on the short-term learning component of biological systems which is a very important component in adaptation to changing environment.

Biological systems operate in time-changing environments, and they are especially good at learning in such conditions. Biological learning appears to be at the basis of the capacity of biological systems to adapt to time-varying environments [10],[11]. The learning components of biological systems may be divided into three types: sensory, short-term and long-term [12]. The sensory component provides immediate simple reactions to certain changes in the environment. One example of the sensory component is human reflex, which is almost instant in response to stimulus. The long-term learning component is based on experiences gained over periods of time. The short-term learning stands in between the sensory and long-term learning components, and is capable of performing more complex learning tasks than the sensory component while being able to adapt to the changing environment quickly.

The interplay among these biological learning components appears to provide to the organism the capacity to operate in uncertain environments. In addition, two cognitive processes that help in preventing overloading are s*ensory gating* (by which the brain adjusts its response to stimuli) and *selective attention* (which allows the brain to concentrate on one aspect to the exclusion of others). This is seen, for example, in the driving of a car, where the driver may not be paying total attention to the road in ordinary circumstances. But when the driving conditions are difficult, a different center takes over the driving completely. The first of the two cases may be ascribed to long-term learning and the second to short-term learning.

In this paper, we propose a hybrid architecture for on-line learning of a time-varying system. The proposed hybrid architecture employs a surface learning component which can adapt to quick changes in system conditions, as well as a deep learning component which is highly accurate to minimize errors within the same regime of operation. A high level cognitive agent monitors the outputs from the surface learning and deep learning components, and automatically generates desirable system outputs.

We show that the proposed approach provides much better performance compared to traditional approaches. The proposed model provides a framework which may lead to flexible intelligence that is able to adapt to changing environment quickly. It will be helpful for processing of information associated with complex "intelligent" tasks in the real world with dynamic and intentional phenomena in the presence of uncertainty about the environment. Specifically, we will be able to explore novel and hybrid methods that extend the breadth of reasoning to include uncertain and dynamic environments, and also consider the performance with interacting agents.

The rest of the paper is organized as follows. Section 2 described the proposed hybrid system architecture. Section 3 discusses some candidate components in building the hybrid system. Section 4 evaluates the performance of the proposed hybrid system. We conclude our work in Section 5.

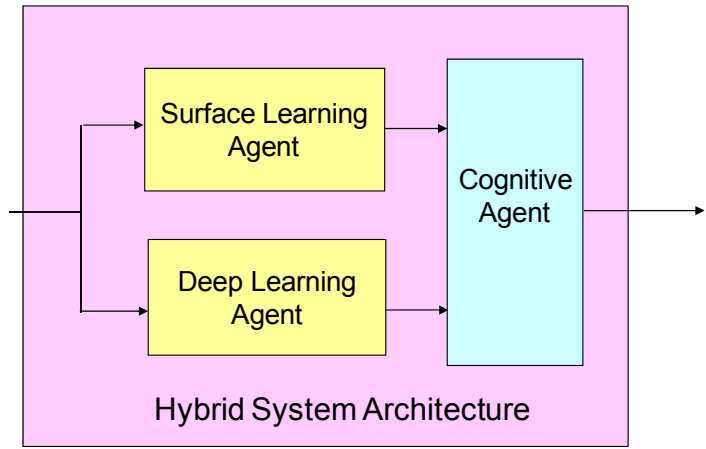

**Fig. 1. Hybrid System Architecture**

## 2   Hybrid System Architecture for On-Line Learning

The proposed hybrid system can be applied to a broad range of applications such as function approximation, time series prediction, and pattern classification. Each of these areas naturally covers many specific applications. For example, time series prediction may be applied to financial or signal data, and pattern recognition includes application to engineering, military, or medical systems. In this section, we describe the general framework of the proposed hybrid system architecture that enables fast and accurate on-line learning of a time-varying system.

Figure 1 shows the proposed hybrid system architecture, which consists of a *surface learning agent*, a *deep learning agent*, and a *cognitive agent*. In the hybrid system architecture, the surface learning agent and the deep learning agent work in parallel in responding to the changes of the environment. In particular, the surface learning agent picks up the short-term structure of the information signal, and performs quick analysis to respond to the stimulus. The deep learning agent seeks out long-term structure of the information, and abstracts longer correlations present within the system. The high level cognitive agent monitors the environment, as well as the decisions made by the surface and deep learning agents, and automatically generates the most desirable system output based on the situation.

Consider the system of Figure 2 where the input $X(t)$ is generated by a random switching between several random processes $\{X_i(t), i = 1, ..., N\}$; $S$ is a dynamic system; and $Y(t)$ is the output random process. The objective is to predict the realization $Y(t)$ of the output process from its past values.

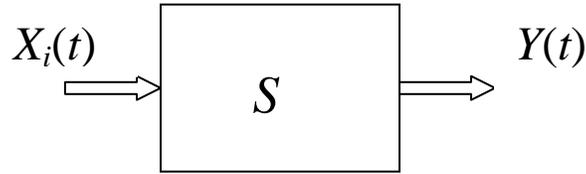

**Fig. 2. System Under Consideration**

The motivation for solving this problem is the consideration of real world processes where the input is time varying in such a manner that it is best represented as a series of different random processes. For an agent that wishes to predict $Y(t)$ based on its past, it is essential to learn which of the $N$ $X_i$s is at the input, where it is assumed that the statistical characteristics of the $X_i$s are known so that its prediction can be made relatively easily. The problem arises from the fact that the input signal switches between different processes and therefore no single statistical technique will work at the output all the time.

We divide the problem of prediction into two components: *deep learning* and *surface learning*. Deep learning operates within the regime of a specific index $i$ of $X_i$ and provides the best performance with respect to some criterion (such as minimization of RMS error). One would expect that it would take the deep learning agent some amount of time to adapt to the statistical characteristics of the signal at the input and, therefore, it would be assumed that the switching between the modes takes place slower than this characteristic time. Surface learning operates best during the transition regime between, let us say, the indices $i$ and $j$.

The dichotomy of the surface and deep learning agents in the proposed hybrid system can find its success in linguistics [13]. For example, in speech or text understanding, without knowing the context, a narrative may appear meaningless. The syntax is provided by the surface structure, whereas the semantics come from the deep structure. Clearly, the surface learning and deep learning agents in the proposed hybrid system is not quite identical to that of syntax and semantics, but it does constitute a step in this direction.

The role of the cognitive agent can be as simple as making a binary decision of selecting the system output from either the surface learning agent or the deep learning agent, or as complex as synthesizing the outputs of the two learning agents into a high level of intelligence. For example, the cognitive agent in the hybrid architecture monitors the errors produced by the surface learning agent and the deep learning agent, and automatically switches the system output according to the system conditions. When the system function changes suddenly, the surface learning agent is able to catch up with the changes quickly while the deep learning agent requires more time to produce results with acceptable outputs. In this case, the cognitive agent will use the outputs produced by the surface learning agent. When the system stabilizes in a certain operational region, the deep learning agent will eventually catch up and produces outputs with more accuracy. When this happens, the cognitive agent will select the outputs from the deep learning agent instead. As a result, the hybrid architecture can respond to the changing system quickly and accurately.

## 3   Building Blocks of Hybrid Architecture

The deep learning agent in the hybrid system must have the property of disregarding quick changes or transients in the input. In other words, it must learn such exemplars that are typical of an established mode. The deep learning agent is, therefore, not very good at determining the transition between modes and this is the reason why it takes more time than an agent that is focused on learning quick changes in the signal. In the biological domain, the deep learning agent is the collection of long-term learning mechanisms.

Several neural networks listed in the section on historical notes of [14] could serve as the deep learning agent. In this paper, we use the multi-layer perceptron with back-propagation as the deep learning agent because it is able to learn the long-term characteristics of the patterns most efficiently [15]. For the rest of the section, we focus on the choices of surface learning agent, which is the most challenging component in designing the hybrid system.

The surface learning agent needs to be able to learn the changed system function instantaneously or near-instantaneously. The performance of the surface learning agent directly affects the performance of the hybrid system. We describe two candidates for the surface learning agent in detail as follows.

### 3.1 Corner Classification (CC) Networks

The *corner classification* (CC) network is based on the idea of phonological loop and the visio-spatial sketchpad. It was proposed by Kak in three variations [8],[9]. These and its more advanced variants are also known as the class of Kak neural networks. There are four versions of the CC technique, represented by CC1 through CC4. The concept of radius of generalization was introduced in CC3 and thus this neural network overcame the generalization problem that plagued the earlier CC2 network. Each node in the network acts as a filter for the training sample. The filter is realized by making it act as a hyper plane to separate the corner of the $n$-dimensional cube represented by the training vector and hence the name corner-classification technique. The CC4 is shown to be better than the other networks in the CC category [16], and is described in detail as follows.

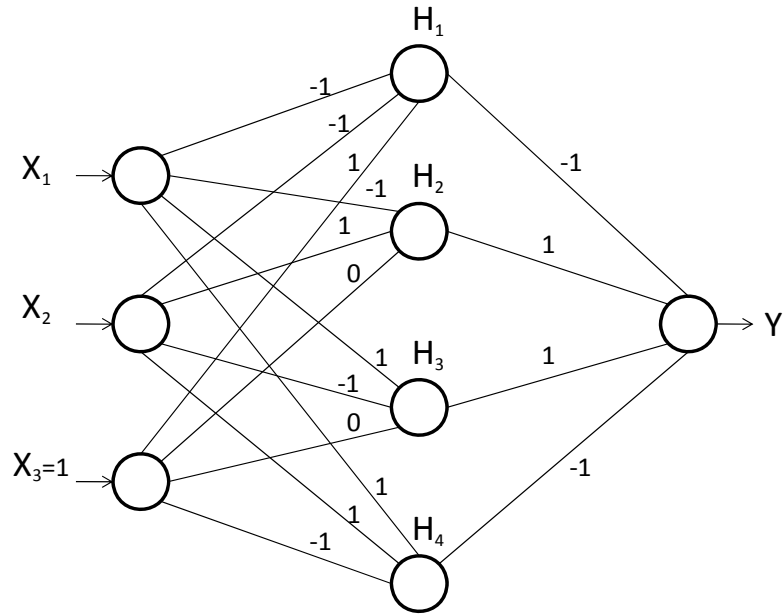

**Fig. 3. CC4 network architecture for learning the XOR function**

CC4 uses a feedforward network architecture consisting of three layers of neurons as shown in Figure 3. The number of input neurons is equal to the length of input patterns or vectors plus one, the additional neuron being the bias neuron, which has a constant input of 1. The number of hidden neurons is equal to the number of training samples, each hidden neuron correspond to one training example. The last node of the input layer is set to one to act as a bias to the hidden layer. The binary step function is used as the activation function for both the hidden and output neurons. The output of the activation function is 1 if summation is positive and zero otherwise.

Input and output weights are determined as follows. For each training vector presented to the network, if an input neuron receives a 1, its weight to the hidden neuron corresponding to this training vector is set to I. Otherwise, it is set to -1. The bias neuron is treated differently. If $s$ is the number of I's in the training vector, excluding the bias input, and the desired radius of generalization is $r$, then the weight between the bias neuron and the hidden neuron corresponding to this training vector is $r - s + 1$. Thus, for any training vector $x_i$ of length $n$ including the bias, the input layer weights are assigned according to the following equation:

$$w_i[j] = \begin{cases} 1 & \text{if } x_i[j] = 1 \\ -1 & \text{if } x_i[j] = 0 \\ r - s + 1 & \text{if } j = n. \end{cases}$$

The weights in the output layer are equal to 1 if the output value is 1 and −1 if the output value is 0. This amounts to learning both the input class and its complement and thus instantaneous. The radius of generalization, $r$ can be seen by considering the all-zero input vectors for which $w_{n+1} = r + 1$. The choice of $r$ will depend on the nature of generalization sought. Figure 3 presents a network for the implementation of the XOR example using CC4. Since the weights are 1, -1, or 0, it is clear that actual computations are minimal. In the general case, the only weight that can be greater in magnitude than 1 is the one associated with the bias neuron.

### 3.2 Fast Classification (FC) Network

Here we present another candidate for the surface learning agent, the fast classification (FC) network [17][18], which is a generalization of the CC networks. Whereas the CC networks perform nearest-neighbor generalization of data as binary vectors, the FC network does the same by considering analog-valued vectors. The FC network uses some appropriate metric to compare the stored vectors of the data to the input data. It depends on the concept of "nearest neighbor" for generalization and consists of three layers - an input layer, a hidden layer and an output layer.

The input data is normalized and presented as input vector $\boldsymbol{x}$. The hidden neuron is represented by the weight vector $\boldsymbol{w_i}$ and its elements are represented by $w_{i,j}$, $i = (1,2,\ldots,S)$ and $j=(1,2,\ldots,R)$, where $R$ is the number of components of the input vector and $S$ is the number of hidden neurons (the number of training samples). The output is the dot product of the vectors $\boldsymbol{\mu}$ and $\boldsymbol{u}$, where $\boldsymbol{\mu}$ is the vector at the output of the Rule Base and $\boldsymbol{u}$ is the vector of weights in the output layer as shown in Figure 4. This network can be trained with just two passes of the samples, the first pass assigns the synaptic weights and the second pass determines the radius of generalization for each training sample by fuzzification of the location of the each training sample and by assigning fuzzy membership functions of output classes to new input vectors.

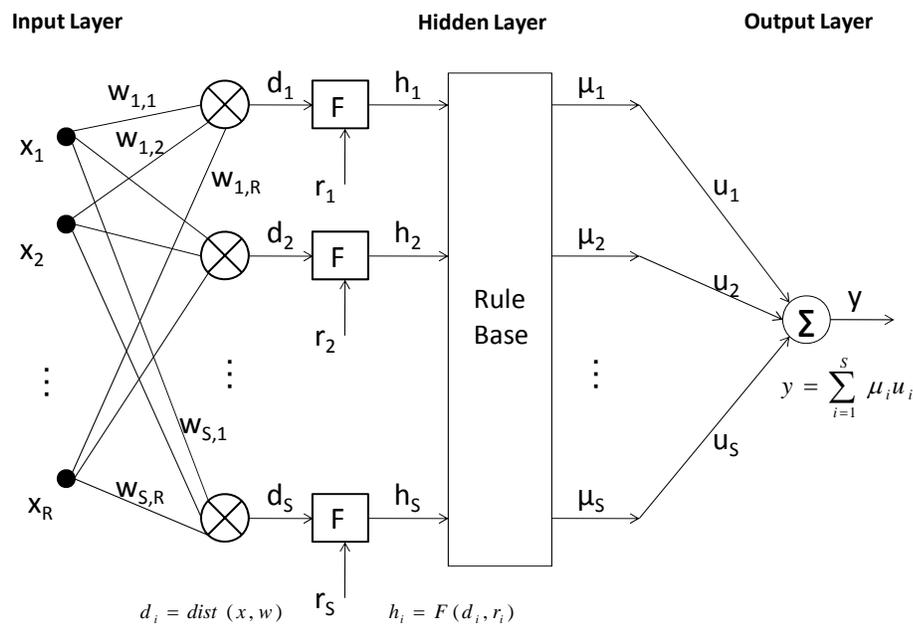

**Fig 4: FC network architecture**

The network behaves as a 1NN classifier and a kNN classifier (1- or k-nearest neighbor classifier) according to whether the input vector falls within the radius of generalization of a training vector or not. The radius of generalization acts as a switch between the 1NN classifier and the kNN classifier. The FC network meets the specifications set by traditional function approximation that every data point is covered in the given training sample and also Cover's theorem on separability of patterns [19]. In the practical case, the *k* values are determined by the sample size and be a fraction of the sample size. If k=S then the FC network operating as a kNN classifier can be viewed as a RBF network [14] provided the membership function is chosen to be a Gaussian distributed. On the other hand, if the weighting function is chosen to be the membership function, the FC classifier can be considered as kernel regression [18]. As in the CC networks, this network requires as many hidden neurons as the number of training samples (although the number of hidden neurons could be trimmed to a certain extent).

As mentioned before, the CC networks work on binary data, which requires quantization of input values. If that is a constraint one wishes to avoid for certain reasons, one can use FC networks, which are a generalization of CC networks and they work on continuous valued data. FC networks require a comparison with stored information in one pass, and, therefore, they are not quite instantaneous, but the calculation can be done in time that could be smaller than the time instant at which the next data comes in.

## 4. Performance Evaluation

In this section, we evaluate the performance of the hybrid system using several benchmarks.

### 4.1 Mackey-Glass Time Series Prediction

Time series prediction is widely used in financial data for prediction of stocks, currency, interest rates and engineering such as electric load demand. The network is trained by historical data set with time index and the network predicts the future values based on past values. Mackey-Glass (MG) time series is a chaotic time series based on Mackey-Class differential equation,

$$\frac{dx(t)}{dt} = -B * x(t) + a * \frac{x(t-D)}{1 + x(t-D)^C},$$

where the choice of *D* has important influence on its chaotic behavior. In most research work, *D* is set to be 17 or 30 because the interesting quasi-periodic behaviors these two values present.

We investigate the system response of the hybrid system when the MG series migrates from parameter set 1 to parameter set 2. More specifically, the hybrid system is first trained to predict MG series with parameter set 1 (**P1** = {A = 0.8, B=0.1, C=10, D=30}), and then the MG series with parameter set 2 (**P2** = {A=0.8, B=0.1, C=10, D=17}). The sliding window is of size 6, and thus inputs consist of 6 consecutive points that are use to predict the very next point ahead: ( $\hat{x(k+1)} = f\{x(k), x(k-1),..., x(k-5)\}$ ).

We obtained the raw data from the Working Group on Data Modeling Benchmarks (under IEEE Neural Networks Council Standards Committee). The first 3500 points in the initialization stage are not used for either training or testing. The next 1000 points are used for training and validation. Another 500 points are used for testing.

The performance of CC and FC are compared as the potential candidates for the surface learning agent. The multi-layer perceptron with back-propagation (BP) is used for the deep learning agent. We are interested in studying how CC, FC and BP will respond to the change of parameters in the MG series. We use the RMS error as the performance measure of the system response to the relearning process. In Figure 5, we observe that as the candidates for the surface learning agent, CC and FC has reduced RMS error almost instantaneously. On the other hand, it takes some time for the deep learning agent BP to converge to new parameters. However, in a long run, the RMS error produced by the deep learning agent BP is smaller than the surface learning agent CC or FC. The vertical line marks the stopping point (at 1886-th adaptation cycle) which is determined by the RMS error validation set so as to avoid system over-fitting.

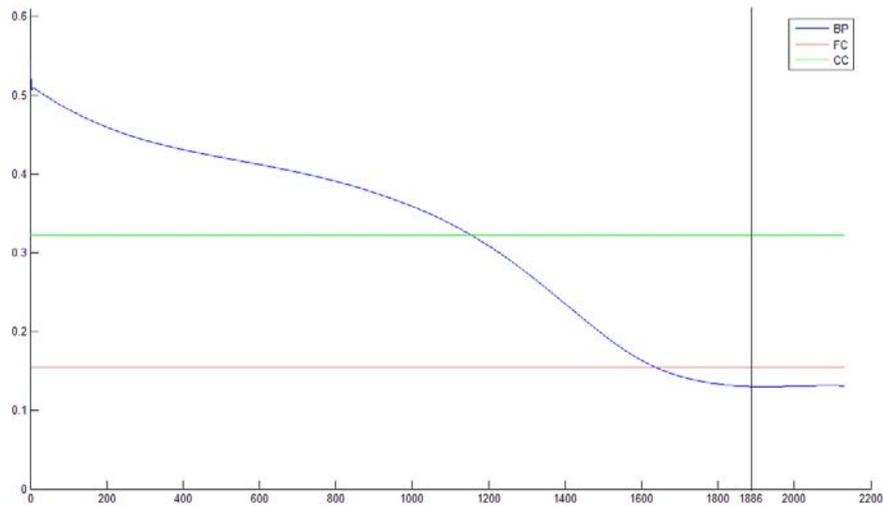

**Fig. 5  RMS Errors vs. Adaptation cycles**

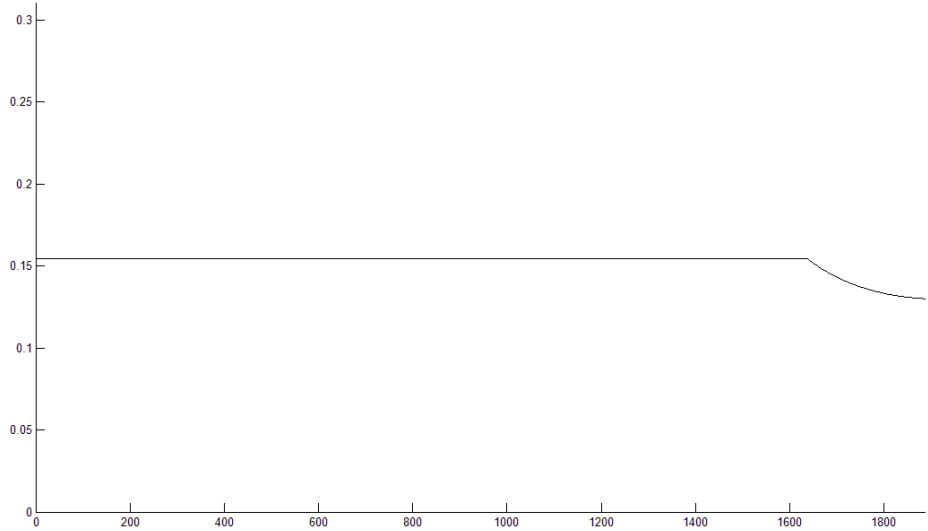

**Fig. 6 RMS error of the hybrid system**

Since FC produces smaller RMS errors than CC, we use FC and BP as the surface learning agent and the deep learning agent in the hybrid system, respectively. The cognitive agent automatically selects the output from the surface learning agent FC once it detects that the system function changes, and switches to the deep learning agent BP after it catches up. The RMS error of the hybrid system is shown in Figure 6. It is clear that the hybrid system has enhanced performance, taking advantage of the surface learning agent and the deep learning agent.

### 4.2 CATS Benchmark Test

The CATS benchmark arises from the well-known Time Series Prediction Competition held in 2004, in which participants were invited to predict 100 missing points from a 5000-point artificial time series. This 100 data points were divided into 5 groups, each located in the position range 981-1000, 1981-2000, 2981-3000, 3981-4000 and 4981-5000, respectively. Participants proposed several prediction methods, linear as well as nonlinear and the winner suggests that good division of the problem into short-term prediction and long-term prediction lead to good result. Since we are interested in the system response of the hybrid system, we skip the input selection stage and only the use short-term prediction. The short-term prediction is known as $\hat{x}(t+1) = f(x(t), x(t-1),..., x(t-W+1))$, where $W$ indicates the window size.

Each time after a prediction is made, the oldest point in the window will be dropped and the newly predicted point is appended on the other side, becoming the current value to predict the next point. We use validation sets of 10 points (from the known points) to find the best window size, with which we train the final network. We first train the networks to learn points 1-980 and 1001-1980 to predict points 981-1000. Thereafter, based on the trained network, relearning is launched to learn another two segments 3001-3980 and 4001-4980. The system response during the relearning process is thus evaluated. In both cases, 10-point validation sets are utilized, and they are selected to be points 971-980 and 3971-3980, which are adjacent to the to-be-predicted points 981-1000 and 3981-4000, respectively. CC precision is fixed at 256 and window size is selected to be 30 based on the results of validation sets.

Figure 7 plots the $E_1$ errors for points 3981-4000 ($\dfrac{\sum\limits_{t=3981}^{4000}(y_t - \hat{y_t})^2}{100}$) produced by BP, CC and FC during the relearning process. The BP result is obtained by averaging the results of 10 different trials. Note that the average stopping point for BP is at the 502th adaptation cycle determined by the validation set. As expected, FC produces good results almost instantly. BP converges slowly but achieves better results eventually. The large error produced by CC is due to the precision. Since FC is consistently better than CC, we use FC as the surface learning agent and BP as the deep learning agent for the hybrid system. The hybrid system switches automatically between the surface learning agent FC and the deep learning agent BP in the relearning process. The $E_1$ error of the hybrid system is shown in Figure 8. This further demonstrates the advantages of the hybrid system.

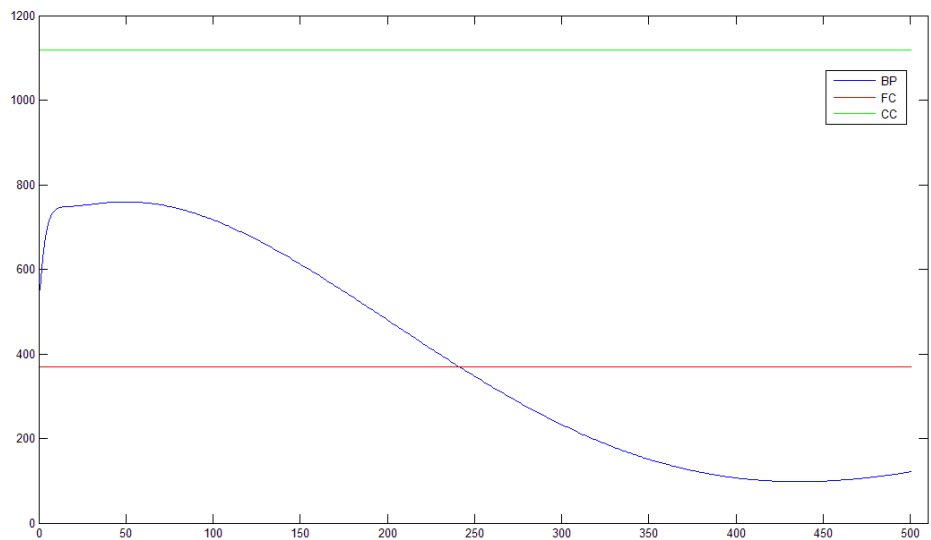

**Fig. 7 $E_1$ Errors during the relearning process**

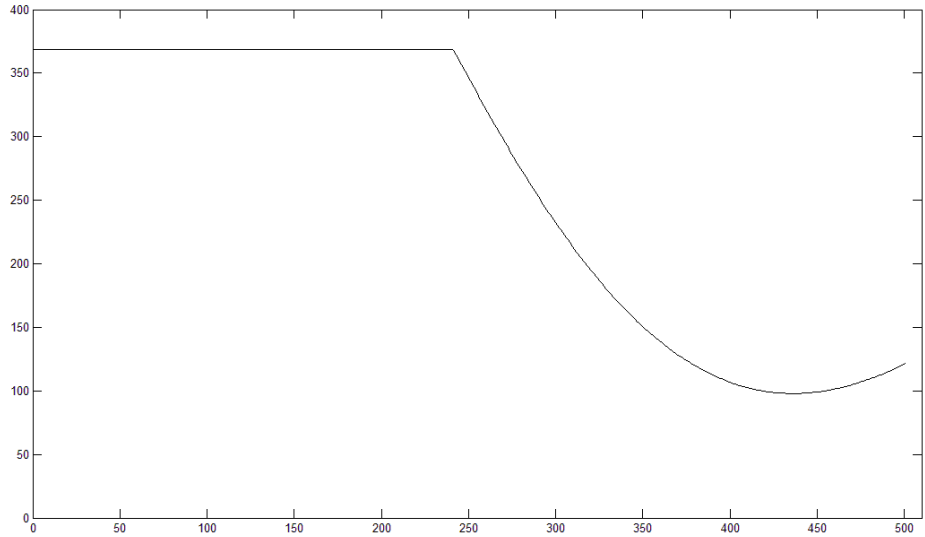

**Fig. 8  E₁ Error produced by the hybrid system**

### 4.3 Smooth Function Approximation

The problem in smooth function approximation is to implement a function, $F(x)$, which approximates an unknown smooth function, $f(x)$, with a priori as the input-output pairs in a Euclidian sense for all inputs.

Again we use the CC and FC as the potential candidates for the surface learning agent, and BP as the deep learning agent. The hybrid system is first trained to approximate a smooth function with parameter set P1, and then relearns the function as the parameter set changes to P2. The objective smooth function is selected to be the pdf of the Multivariate Normal Distribution, $f_X(x_1, x_2) = \frac{1}{2\pi |\Sigma|^{1/2}} \exp(-\frac{1}{2}(x-\mu)^T \Sigma^{-1}(x-\mu))$ . The configurations of the two parameter sets are in shown table I. The training set is constructed by uniformly sampling 900 points in the 2-D input space, and another 1600 points make up the testing set.

| P1 | P2 |
|---|---|
| μ={0,0} | μ={-0.2,0.7} |
| ∑={0.8 0.2;0.2 0.1} | ∑={0.25 0.3;0.3 1} |

**Table I. Smooth function parameter configuration**

To understand how the deep learning agent and the surface learning agent in the hybrid system respond to the changed function, we plot a series of transitional plots as follows. The actual function with parameter set P1 was plotted in Figure 9. Figure 10-12 capture the relearning processes for the deep learning agent BP. It can be observed that the deep learning agent BP adapts from parameter set P1 to P2 gradually. Figure 13 and Figure 14 show the approximation of the surface learning agent candidates CC and FC after instantaneous training, respectively.

The RMS errors produced by the BP, CC and FC during the relearning processes are plotted in Figure 15. As we can see from the figure, it takes a considerable number of adaptation cycles for the deep learning agent BP to catch up with the surface learning agent CC or FC. Since FC produces smaller RMS errors, we choose FC as the surface learning agent for smooth function approximation. The RMS errors of the hybrid system are plotted in Figure16.

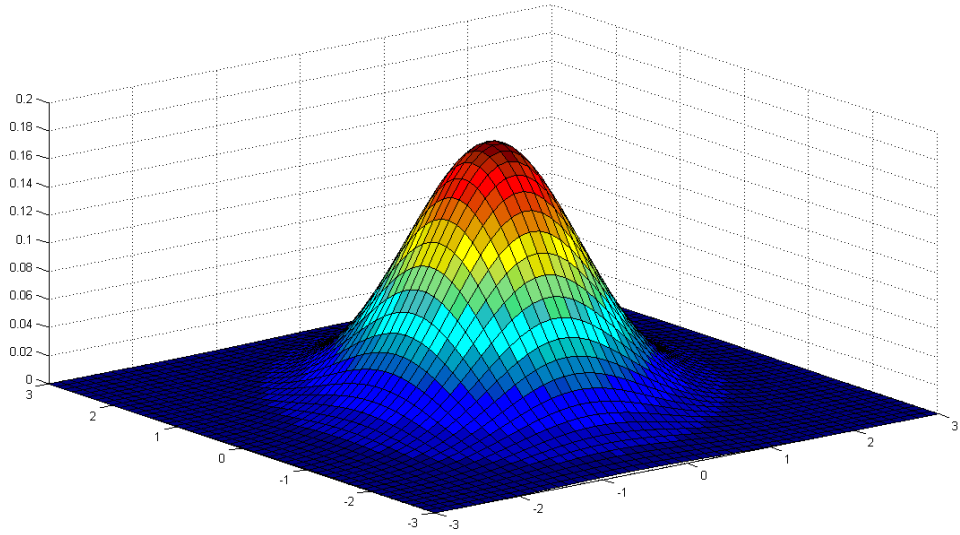

**Fig. 9 Actual smooth function with parameter P1**

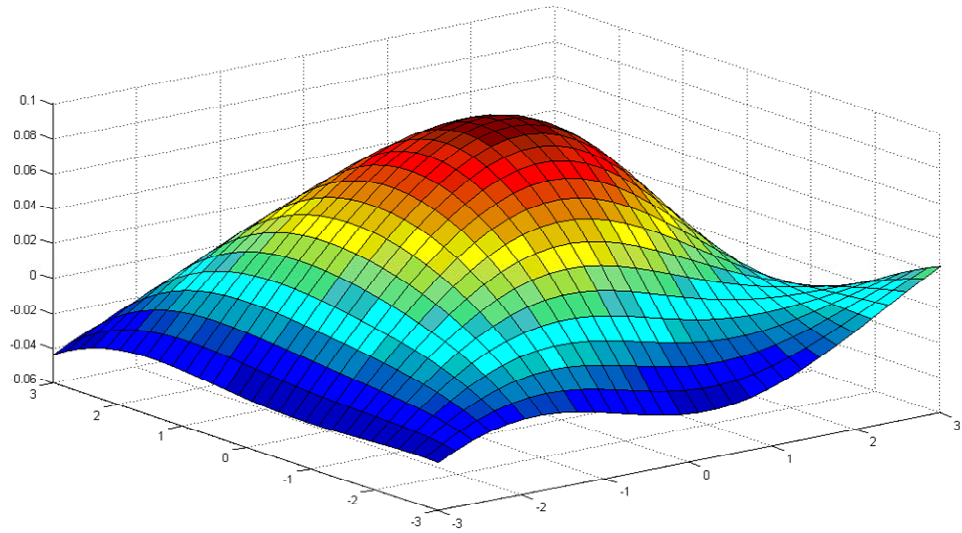

**Fig. 10 BP shape after 1 adaptation cycles**

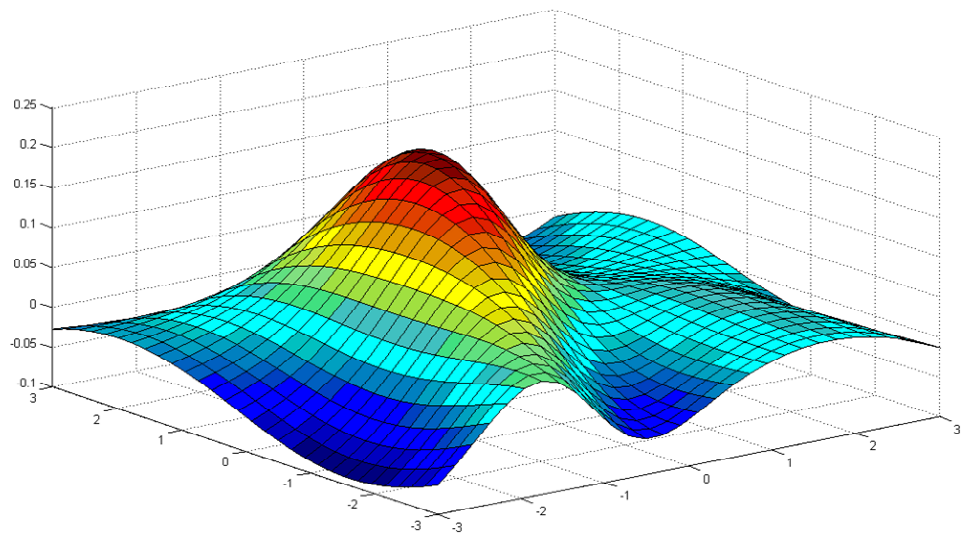

**Fig. 11 BP shape after 5001 relearning adaptation cycles**

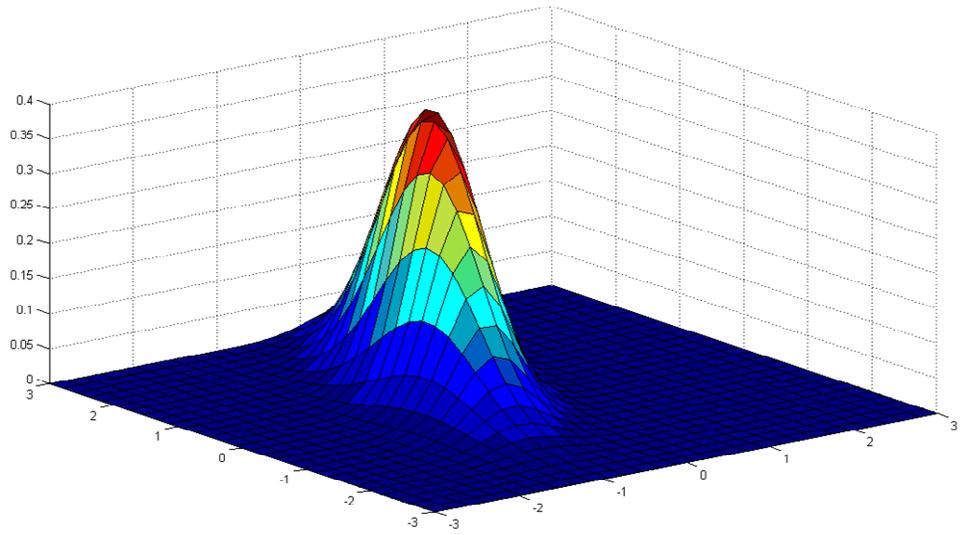

**Fig. 12 BP shape after 300003 relearning adaptation cycles**

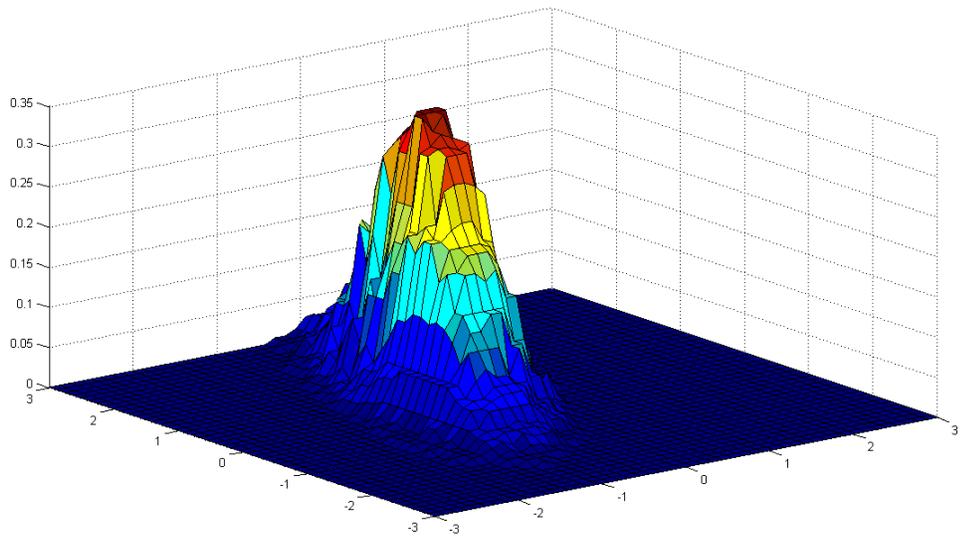

**Fig. 13 CC Approximation of P2**

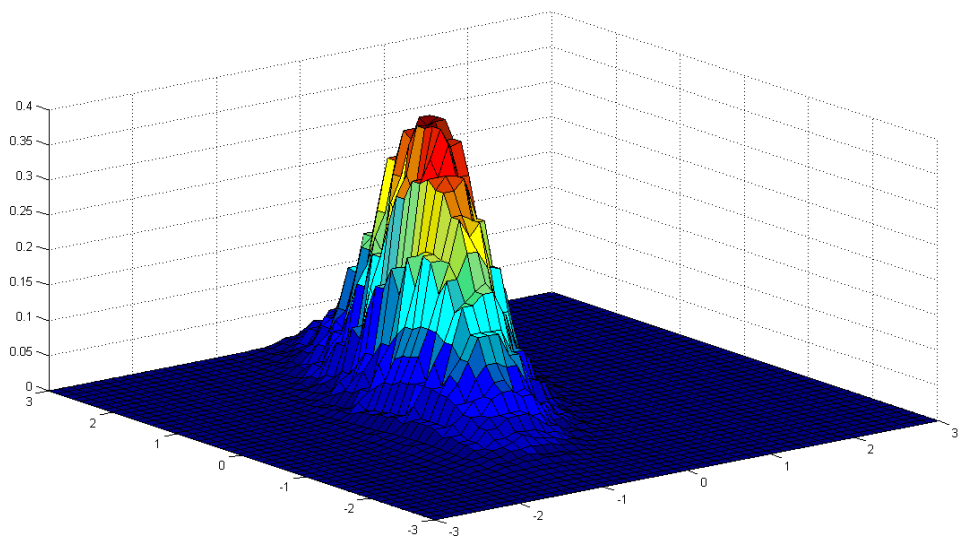

**Fig. 14 FC approximation of P2**

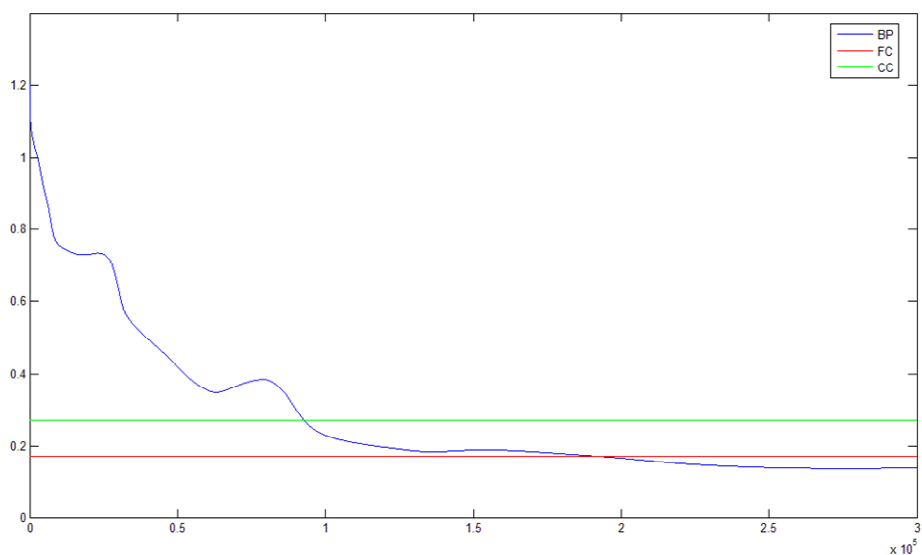

**Fig. 15 RMS errors vs. Adaptation cycles**

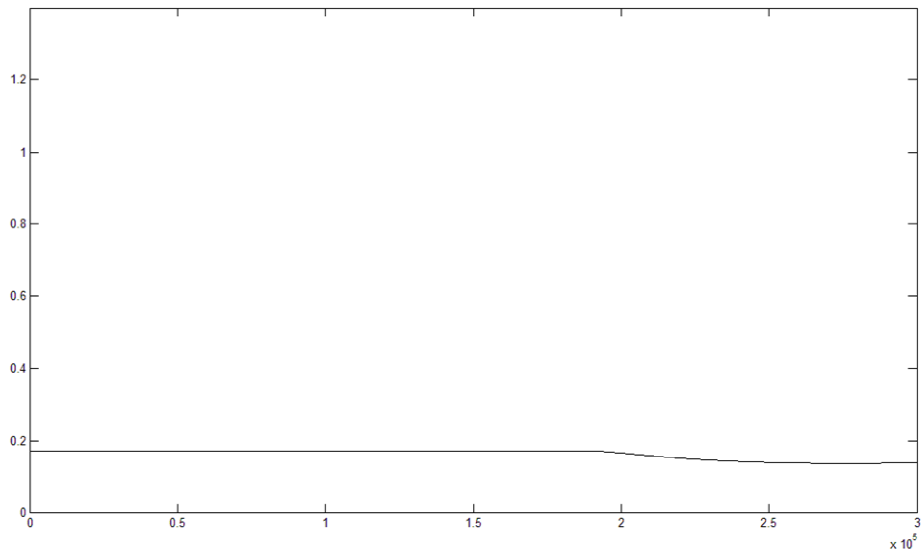

**Fig. 16 RMS of the Hybrid Architecture**

## 5. Conclusion

This paper has proposed a hybrid neural network architecture that uses two kinds of neural networks simultaneously: (i) a surface learning network that can quickly adapt to new modes of operation; and, (ii) a deep learning, accurate network for use within a specific regime of operation. The two networks of the hybrid architecture perform functions like that of short-term and long-term learning. Such a hybrid architecture can be based on different classes of learning systems although in this paper we have considered CC and FC fast learning networks on the one hand and back-propagation networks on the other.

The performance of the hybrid architecture has been compared with that of multi-layer perceptron with back-propagation and the CC and FC networks for chaotic time-series prediction, the CATS benchmark test, and smooth function approximation. It has been shown that the hybrid architecture provides a superior performance based on a RMS error criterion.

We expect the applications of our hybrid neural network architecture to problems in finance, control, time-series prediction, and economic forecasting.